\documentclass[conference]{IEEEtran}
\IEEEoverridecommandlockouts
\usepackage{cite}
\usepackage{amsmath,amssymb,amsfonts}
\usepackage{algorithmic}
\usepackage{graphicx}
\usepackage{textcomp}
\usepackage{booktabs}
\usepackage{subcaption}
\usepackage[hidelinks]{hyperref}
\usepackage{multirow}
\usepackage{xcolor}
\def\BibTeX{{\rm B\kern-.05em{\sc i\kern-.025em b}\kern-.08em
    T\kern-.1667em\lower.7ex\hbox{E}\kern-.125emX}}

\usepackage{comment}

\begin{document}

\IEEEoverridecommandlockouts
% \IEEEpubid{\begin{minipage}[t]{\textwidth}\ \\[10pt]
% \normalsize{979-8-3195-4163-5/26/\$31.00 \copyright2026 IEEE}
% \end{minipage}}
\title{A Unified Framework for License Plate Identification and Character-Recognition\
%{\footnotesize \textsuperscript{*}Note: Sub-titles are not captured in Xplore and
%should not be used}
% \thanks{Dewant Katare is currently with Perciv AI. The presented work was done during his time with TU Delft, The Netherlands.}
}

\author{\IEEEauthorblockN{Saurabh Kumar}
\IEEEauthorblockA{\textit{Walmart} \\
%\textit{}\\
Sunnyvale, USA \\
saurabh.kumar4@walmart.com}
\and 
\IEEEauthorblockN{Dewant Katare}
\IEEEauthorblockA{\textit{Delft University of Technology} \\
%\textit{}\\
Delft, the Netherlands \\
d.katare@tudelft.nl}
\and 
\IEEEauthorblockN{Prakriti Jain }
\IEEEauthorblockA{\textit{Heidelberg University} \\
%\textit{}\\
Heidelberg, Germany \\
prakriti.jain@stud.uni-heidelberg.de}
}

% \author{
%   \IEEEauthorblockN{Anonymous Author(s)}
%   % \IEEEauthorblockA{\textit{Affiliation(s) Omitted for Blind Review} \\
%   Paper ID: \textit{[51]}}
% }

\maketitle

\begin{abstract}
Automatic License Plate Recognition (ALPR) is used in traffic monitoring, smart parking, toll collection, and vehicle access systems, where license plates must be localized and their alphanumeric content are recognized from continuous video streams. Most ALPR systems use separate detection and recognition networks, which increases repeated feature extraction, memory use, and inference latency, especially on low-power edge devices. This paper proposes a unified framework for license plate detection and character recognition on edge hardware. The proposed network uses a shared CSP-Darknet and PANet feature encoder with an anchor-free localization head and a RoIAlign-based sequence-recognition branch. Character decoding is performed directly on aligned feature maps, avoiding image-space plate cropping and a second recognition backbone. The model is trained on an AI-assisted dataset of approximately 37,000 manually verified images covering license plate layouts from the United States, with variations in illumination, viewpoint, weather, blur, occlusion, and plate design. The trained model is converted to an INT8 TensorRT engine and evaluated on a Jetson Orin Nano with 8~GB memory. The INT8 model achieves an $\mathrm{mAP}_{50}$ of 97.8\% and a full-sequence accuracy of 90.7\%, with 147~FPS throughput, 6.8~ms end-to-end latency, 315~MB allocated inference memory, and 9.4~W average power consumption. Relative to the YOLOv8s+CRNN baseline, the proposed deployment provides a 7.1$\times$ throughput increase and a 77.8\% reduction in allocated inference memory.
\end{abstract}


\section{Introduction}
\label{sec:introduction}

Automatic License Plate Recognition (ALPR) supports intelligent transportation, toll collection, smart parking, traffic monitoring, fleet management, and public safety. These applications require accurate processing of continuous video under daytime and nighttime illumination, motion blur, rain, occlusion, and viewpoint changes. Such requirements become harder to meet on embedded platforms with limited computing resources, memory, and power.

Most ALPR systems use a two-stage pipeline in which a detector first localizes the plate and an independent recognition model predicts its character sequence~\cite{zou2022license,pan2022hybrid,ismail2025dual,reddy2024license}. Although this design can provide high recognition accuracy, executing two networks increases latency, memory use, and computational cost. Unified architectures instead share feature extraction between localization and recognition, allowing both tasks to run within one forward pass~\cite{qin2022towards,gao2024toward}. However, the effect of INT8 quantization on shared features remains insufficiently studied because localization and character decoding may respond differently to reduced numerical precision.

Edge processing reduces communication delay and removes the need for continuous cloud connectivity. Prior work has examined resource allocation and energy-aware execution for embedded intelligence~\cite{katare2023energy,katare2024arasec,katare2025energy}, while lightweight ALPR models have targeted embedded processors and microcontrollers~\cite{tang2024ultra,sonnara2025efficient}. A further challenge is the visual variation among the license plates, including differences in typography, background artwork, spacing, state labels, and reflective materials. Existing datasets and models often cover limited regional formats or require additional processing stages to address such variation~\cite{quraishi2024detection,usama2025vehicle,sultan2023towards}.

This work addresses two gaps: the effect of INT8 quantization on shared localization and recognition features, and the ability of a single-pass model to process diverse United States plate layouts under edge latency and memory limits. Accordingly, the following research questions are examined:

\begin{itemize}
    \item \textbf{RQ1:} How does INT8 quantization affect localization and sequence recognition in a shared-backbone ALPR network?

    \item \textbf{RQ2:} How much computation, memory, and latency can feature reuse save relative to dual-stage ALPR while retaining recognition accuracy across diverse United States plate layouts?
\end{itemize}

To answer these questions, this paper makes the following contributions:

\begin{enumerate}
    \item A unified ALPR network that combines a shared feature backbone, an anchor-free localization head, and a RoIAlign-based sequence-recognition branch, removing the second feature encoder used in dual-stage pipelines.

    \item An AI-assisted dataset of approximately 37,000 manually verified images covering license plate layouts from the United States, with synchronized bounding-box and alphanumeric sequence annotations.

    \item A TensorRT INT8 deployment and evaluation on the Jetson Orin Nano 8GB. The optimised model shows $\mathrm{mAP}_{50}$ of 97.8\%, an accuracy of 90.7\%, 147~FPS throughput, 6.8~ms end-to-end latency, 315~MB inference memory, and 9.4~W average power consumption.
\end{enumerate}

\begin{figure*}[t]
\centering
\includegraphics[width=\textwidth]{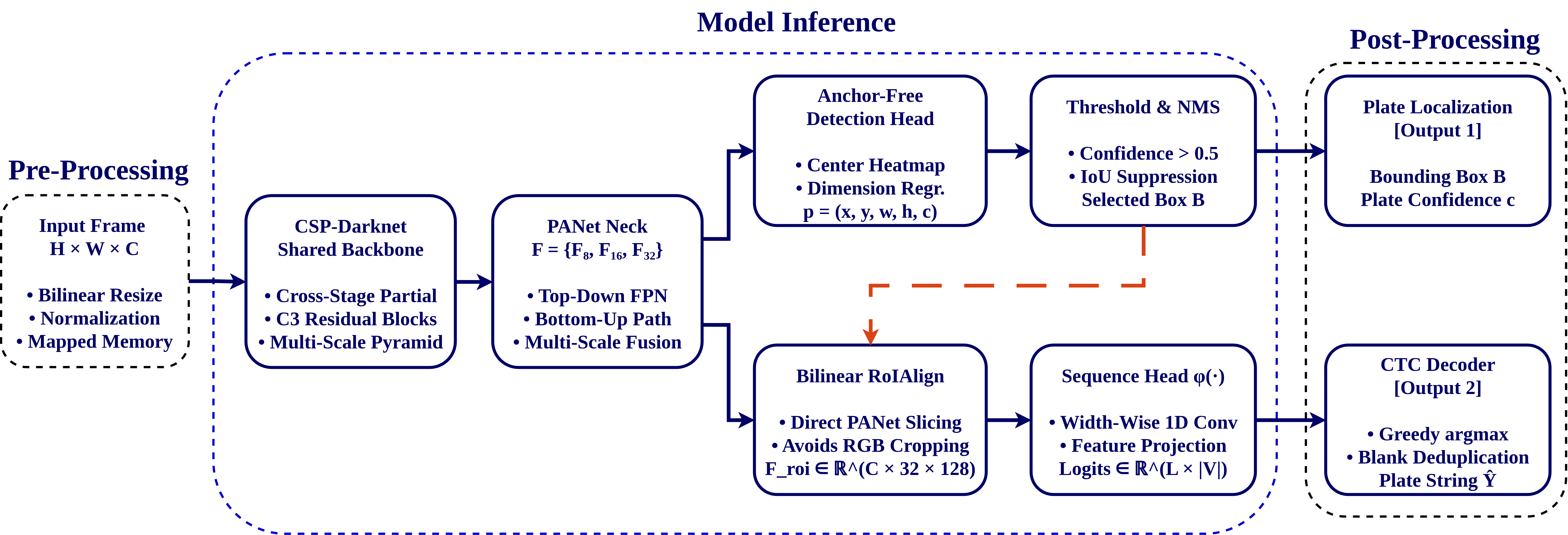}
\caption{End-to-end architecture and runtime execution pipeline of the proposed unified ALPR framework, organized into pre-processing, model inference, and post-processing stages. Bilinear RoIAlign samples plate features directly from the shared PANet pyramid $\mathcal{F}$ using bounding box coordinates $B$ generated by the anchor-free localization head, eliminating duplicate backbone evaluations and image-space cropping.}
\label{fig:pipeline_architecture}
\end{figure*}

\section{Related Work and Challenges}
\label{sec:related_work}

Existing ALPR methods can be grouped into two-stage pipelines, unified end-to-end networks, and edge-oriented systems. These approaches differ mainly in how localization and recognition are connected and how efficiently they run on embedded hardware.

\subsection{Two-Stage ALPR Systems}

Two-stage ALPR systems localize the license plate first and then pass the cropped region to an independent OCR or sequence-recognition model. This separation allows each task to be trained independently, but it requires repeated feature extraction and additional memory transfers. YOLO-based detectors have been paired with ILPRNet, CRNN-style recognizers, and conventional OCR engines for vehicle communication, surveillance, and traffic applications~\cite{zou2022license,pan2022hybrid,ismail2025dual,reddy2024license,anilkumar2024automated}. Although these methods can provide strong recognition accuracy, executing separate detection and recognition networks increases latency, memory use, and computing cost on low-power devices.

\subsection{Unified End-to-End ALPR}

Unified models share visual features between plate localization and character decoding. Qin and Liu~\cite{qin2022towards} proposed a joint localization and recognition network for unconstrained scenes, while Gao \textit{et al.}~\cite{gao2024toward} studied unified processing under variable image resolutions. Shared feature extraction reduces duplicate computation and supports single-pass inference. However, prior studies provide limited analysis of INT8 quantization, shared-feature behavior, and deployment under strict memory and power limits. Their datasets also often cover restricted regional formats, leaving generalization across varied United States plate layouts less studied.

\subsection{Edge-Oriented ALPR}

Recent work has examined lightweight ALPR for embedded processors, microcontrollers, and edge devices~\cite{tang2024ultra,sonnara2025efficient}. Other studies have focused on detector refinement and application-specific systems for smart parking, toll collection, regional surveillance, and OpenCV-based recognition~\cite{chung2024yolo,ditta2025number,usama2025vehicle,derrouz2025mlpr,chandrika2024number}. Hardware-aware execution, adaptive resource management, and energy-aware scheduling have also been studied for edge intelligence~\cite{katare2023energy,katare2024arasec,katare2025energy}. However, many ALPR implementations still use multiple processing stages or optimize model design and deployment separately. The proposed framework addresses these limitations by combining shared feature extraction, RoIAlign-based sequence recognition, and TensorRT INT8 deployment within one pipeline. It is evaluated on a manually verified AI-assisted dataset covering license plate layouts from the United States.

\section{Components and Framework Overview}
\label{sec:framework}

The proposed framework performs license plate localization and character recognition through a unified pipeline designed for real-time edge execution. As shown in Fig.~\ref{fig:pipeline_architecture}, the framework combines shared multi-task inference, sequence decoding, and TensorRT deployment. During inference, each frame is encoded once, and the resulting feature maps are reused by both prediction branches. Plate regions are extracted directly from the feature pyramid using Region-of-Interest Alignment (RoIAlign), avoiding RGB crop reconstruction and a separate OCR encoder. The complete network is converted to an INT8 TensorRT engine for execution on the Jetson Orin Nano.

\subsection{Unified Multi-Task Network}
\label{subsec:multitask_network}

The network consists of a CSP-Darknet backbone, a PANet feature pyramid, an anchor-free localization head, and a RoIAlign-based sequence-recognition branch. These components form a single computational graph in which localization and text decoding share the same visual features.

\subsubsection{Shared Backbone and Localization}

Given an input image $I\in\mathbb{R}^{H\times W\times3}$, CSP-Darknet generates multi-scale feature maps,

\begin{equation}
\mathcal{F}=\{F_{8},F_{16},F_{32}\},
\label{eq:feature_pyramid}
\end{equation}

which are fused by PANet to combine fine spatial detail with higher-level plate information. The anchor-free localization head predicts

\begin{equation}
\mathbf{p}=(x,y,w,h,c),
\label{eq:localization_output}
\end{equation}

where $(x,y)$ is the plate center, $(w,h)$ gives its dimensions, and $c$ is the confidence score. The predicted plate region is passed directly to the recognition branch without forming an intermediate image crop.

\subsubsection{RoIAlign-Based Sequence Recognition}

Conventional two-stage systems crop the detected plate from the RGB image and process it using a second recognition network. In the proposed framework, RoIAlign samples the predicted region directly from the shared PANet features:

\begin{equation}
F_{\mathrm{roi}}
=
\mathrm{RoIAlign}(\mathcal{F},B),
\qquad
B=(x,y,w,h),
\label{eq:roi_align}
\end{equation}

where $F_{\mathrm{roi}}\in\mathbb{R}^{C\times H_r\times W_r}$ is resized to $32\times128$ through bilinear sampling. The aligned tensor is projected into a width-wise sequence and decoded as

\begin{equation}
\hat{Y}
=
\mathrm{CTC}
\left(
\phi(F_{\mathrm{roi}})
\right),
\label{eq:sequence_decoder}
\end{equation}

where $\phi(\cdot)$ is the sequence projection network and $\hat{Y}$ is the predicted plate string. CTC requires only the complete transcription during training, removing the need for character-level bounding boxes.

Feature-space extraction avoids repeated image preprocessing, intermediate RGB allocation, and a second backbone pass. If $B_c$, $H_{\mathrm{loc}}$, and $H_{\mathrm{rec}}$ denote the costs of the backbone, localization head, and recognition head, respectively, the computational costs are

\begin{equation}
\mathcal{C}_{\mathrm{dual}}
=
2B_c+H_{\mathrm{loc}}+H_{\mathrm{rec}},
\quad
\mathcal{C}_{\mathrm{ours}}
=
B_c+H_{\mathrm{loc}}+H_{\mathrm{rec}}
\label{eq:complexity_comparison}
\end{equation}

Thus, the proposed design removes one backbone evaluation per frame.

\subsubsection{Joint Multi-Task Optimization}

The network is trained using the joint objective

\begin{equation}
\mathcal{L}_{\mathrm{total}}
=
\lambda_{\mathrm{box}}\mathcal{L}_{\mathrm{box}}
+
\lambda_{\mathrm{conf}}\mathcal{L}_{\mathrm{conf}}
+
\lambda_{\mathrm{ctc}}\mathcal{L}_{\mathrm{ctc}},
\label{eq:joint_loss}
\end{equation}

where $\mathcal{L}_{\mathrm{box}}$, $\mathcal{L}_{\mathrm{conf}}$, and $\mathcal{L}_{\mathrm{ctc}}$ represent localization, confidence, and sequence-recognition losses. Their weights balance both tasks during shared-backbone training.

\begin{figure}[t]
 \centering
    \includegraphics[width=0.99\linewidth]{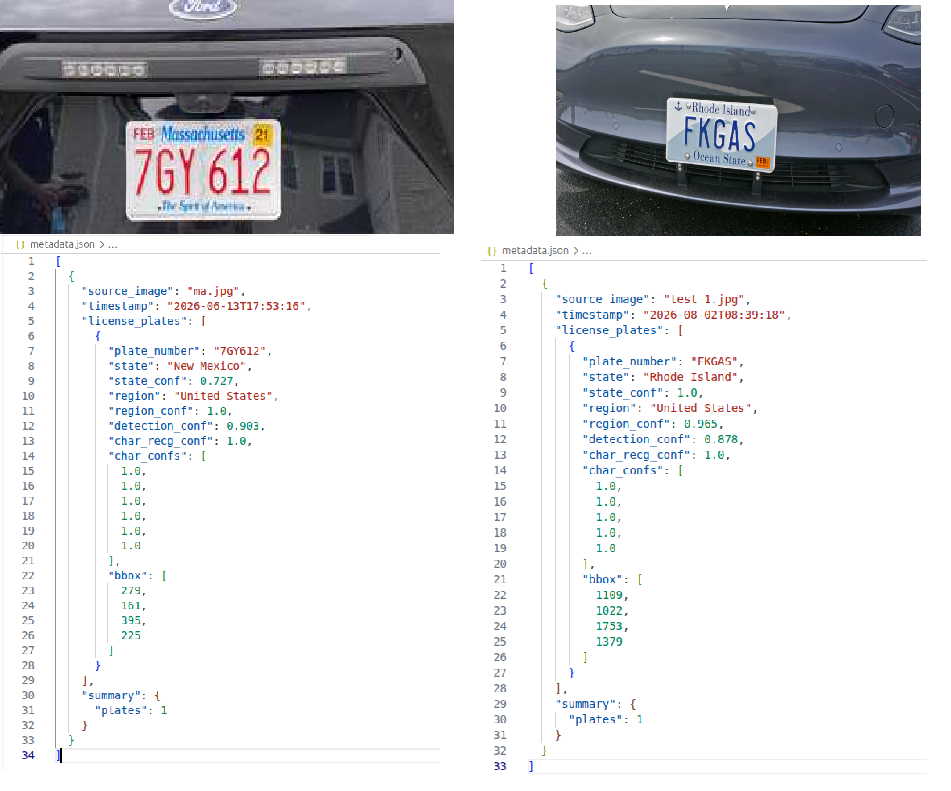}
    \caption{An overview of results and data saved from detection}
    \label{fig:framework}
\end{figure}

Unlike earlier unified ALPR systems focused mainly on end-to-end recognition~\cite{gao2024toward,qin2022towards}, this design performs recognition on RoI-aligned shared features and is structured for TensorRT INT8 execution. This reduces duplicate feature extraction and intermediate memory allocation on embedded hardware. Representative detection outputs and metadata are shown in Fig.~\ref{fig:framework}.

\subsection{AI-Assisted Dataset Generation and Annotation}
\label{subsec:dataset}

A unified ALPR model requires paired localization and transcription labels across varied plate layouts. To provide such data, an AI-assisted pipeline based on Google's Gemini model was used to generate traffic-scene images containing license plates from United States. Prompt templates introduced changes in typography, background design, character spacing, illumination, viewpoint, weather, blur, reflections, and partial occlusion. After quality filtering, the dataset contained approximately 37,000 images.

Each image was manually verified and annotated using a custom interface that combines bounding-box labeling and sequence transcription. Country and state identifiers were stored as metadata, while plate coordinates and complete alphanumeric strings were used for training. Character-level boxes were not required because the recognition branch uses CTC decoding. The dataset was divided using a 70:15:15 ratio, producing 25,900 training images, 5,550 validation images, and 5,550 test images. The split retained samples from all states and includes the primary visual variations introduced during dataset generation.

\subsection{Edge Deployment and Runtime Optimization}
\label{subsec:edge_deployment}

The trained PyTorch model is exported to ONNX and compiled as a single TensorRT engine for execution on the Jetson Orin Nano. The exported graph retains the shared backbone, PANet, localization head, RoIAlign operation, and sequence-recognition branch, allowing detection and recognition to remain within one runtime graph. The engine is configured for a fixed input resolution and batch size $N=1$.

\subsubsection{INT8 Conversion}

The FP32 network is converted to INT8 precision to reduce arithmetic cost and activation memory. For a floating-point value $x$, quantization is defined as:

\begin{equation}
q=
\mathrm{clip}
\left(
\left\lfloor \frac{x}{s}\right\rceil,
-128,
127
\right),
\label{eq:int8_quantization}
\end{equation}

where $s$ is the scaling factor and $q$ is the INT8 value. Since the backbone supplies features to both localization and recognition, calibration samples must cover plate scale, state layout, illumination, viewpoint, blur, and background variation. Entropy calibration is therefore performed using 1,000 training images, with no validation or test samples included.

During engine construction, TensorRT fuses supported operations, selects kernels for the target GPU, and serializes the resulting execution plan for repeated use during video inference.

\begin{table}[t]
\centering
\caption{Summary of the  dataset.}
\label{tab:dataset_summary}
\begin{tabular}{lc}
\hline
Property & Value \\
\hline
Images & 37,000 \\
States & 50 \\
Labels & Box, state, plate text \\
Split & 70:15:15 \\
\hline
\end{tabular}
\end{table}

\subsubsection{Memory and Streaming Execution}

Input, output, and workspace buffers are allocated during initialization and reused across frames. Decoded images are written to page-locked mapped memory that can be accessed by both the CPU and integrated GPU. Preprocessing writes normalized input directly into the TensorRT input buffer when memory layouts permit, reducing explicit copies and repeated allocation.

Feature-space RoIAlign also avoids the RGB crop, resize, and second backbone pass required by two-stage systems. The aligned feature tensor remains in GPU-accessible memory and is passed directly to the sequence-recognition layers.

For continuous batch-one inference, each frame passes through three stages: frame preparation, TensorRT execution, and output decoding. Bounding boxes are filtered using confidence thresholding and non-maximum suppression, while character probabilities are converted to plate strings through CTC decoding. CUDA streams and events allow independent work from adjacent frames to overlap when buffer dependencies are satisfied. The end-to-end latency for frame $k$ is measured as:

\begin{equation}
t_{\mathrm{e2e}}^{(k)}
=
t_{\mathrm{prep}}^{(k)}
+
t_{\mathrm{trt}}^{(k)}
+
t_{\mathrm{post}}^{(k)},
\label{eq:e2e_latency}
\end{equation}

where the three terms correspond to preprocessing, TensorRT execution, and output decoding.

\subsubsection{Target Platform}

Deployment uses a Jetson Orin Nano with 8~GB LPDDR5 memory, a six-core Arm Cortex-A78AE CPU, and an Ampere GPU with 1,024 CUDA cores and 32 Tensor Cores. The runtime environment uses Ubuntu~22.04, JetPack~6.2, CUDA~12.6, cuDNN~9.3, and TensorRT~10.3. Model development uses Python~3.12 and PyTorch~2.7, while the deployed application loads the serialized TensorRT engine through a native runtime interface. Device power settings are controlled through \texttt{nvpmodel}.

\section{Experimental Evaluation}
\label{sec:experiments}

This section evaluates the proposed unified multi-task framework in terms of localization accuracy, character recognition, numerical precision, inference latency, memory allocation, and power consumption. The evaluation includes comparisons with two-stage and unified ALPR baselines, followed by controlled studies of feature extraction and INT8 calibration.

\begin{table*}[t]
\centering
\caption{Localization and recognition results on the held-out test set.}
\label{tab:accuracy_comparison}
\begin{tabular}{lcccc}
\hline
\textbf{Architecture}
&
\textbf{$\mathrm{mAP}_{50}$ (\%)}
&
\textbf{Precision (\%)}
&
\textbf{Char. Acc. (\%)}
&
\textbf{Seq. Acc. (\%)} \\
\hline
YOLOv8s + CRNN
& 98.2 & 97.6 & 94.1 & 88.4 \\

YOLOv8s + PARSeq
& \textbf{98.4} & \textbf{97.9} & \textbf{96.8} & \textbf{92.1} \\

YOLO-LPR without RoIAlign
& 96.5 & 95.8 & 91.2 & 83.7 \\
\hline
Proposed framework, FP32
& 98.1 & 97.4 & 96.3 & 91.5 \\

Proposed framework, INT8
& 97.8 & 97.1 & 95.8 & 90.7 \\
\hline
\end{tabular}
\end{table*}

\subsection{Experimental Setup}

\subsubsection{Training Environment}

Model training and offline analysis were performed on a computing unit consisting of NVIDIA GeForce RTX~4090 GPU with 24~GB of video memory, an Intel Core i9-13900K processor, and 64~GB of system memory. The training environment used Ubuntu~24.04, Python~3.12, PyTorch~2.7, CUDA~12.6, and cuDNN~9.3.

Edge measurements were collected on the Jetson Orin Nano with 8~GB of shared memory. The device used JetPack~6.2 and TensorRT~10.12. All edge experiments were conducted with a batch size of one under a fixed power profile configured through \texttt{nvpmodel}. Unless otherwise stated, the measured latency includes pre-processing, TensorRT execution, and output decoding.

\subsubsection{Dataset Partitioning}

Experiments used the AI-assisted dataset described in Section~\ref{subsec:dataset}. The dataset contains approximately 37,000 annotated images covering license plate layouts. Each sample includes a plate bounding box and its corresponding alphanumeric sequence.

The dataset was divided into training, validation, and test partitions using a 70:15:15 ratio:

\begin{itemize}
    \item Training set: 25,900 images;
    \item Validation set: 5,550 images;
    \item Test set: 5,550 images.
\end{itemize}

All input images were resized to $640 \times 640$ pixels. The test partition was excluded from model training, hyperparameter selection, and INT8 calibration.

\subsubsection{Training Protocol}

The proposed network was trained end-to-end for 100 epochs using the AdamW optimizer. The optimizer parameters were set to $\beta_1=0.9$ and $\beta_2=0.999$, with an initial learning rate of $1\times10^{-3}$ and a weight decay of $1\times10^{-4}$. A cosine annealing schedule reduced the learning rate to $1\times10^{-5}$ by the final epoch. The training batch size was 32. The task weights in the joint loss were configured as

\begin{equation}
\lambda_{\mathrm{box}}=7.5,\qquad
\lambda_{\mathrm{conf}}=1.0,\qquad
\lambda_{\mathrm{ctc}}=0.5.
\end{equation}

These weights balance localization, plate confidence prediction, and sequence recognition during joint training. Data augmentation included random perspective changes of up to $\pm10^{\circ}$, brightness and contrast adjustments of up to $\pm0.2$, Gaussian blur, and motion-blur simulation.

\subsection{Evaluation Metrics}

\subsubsection{Localization Metrics}

License plate localization is evaluated using Intersection over Union (IoU), precision, recall, and mean average precision at an IoU threshold of 0.5, written as $\mathrm{mAP}_{50}$. For a predicted bounding box $B_{\mathrm{pred}}$ and ground-truth box $B_{\mathrm{gt}}$, IoU is defined as

\begin{equation}
\mathrm{IoU}
=
\frac{
\left|B_{\mathrm{pred}}\cap B_{\mathrm{gt}}\right|
}{
\left|B_{\mathrm{pred}}\cup B_{\mathrm{gt}}\right|
}.
\label{eq:iou_metric}
\end{equation}

Precision and recall are computed as

\begin{equation}
P=\frac{TP}{TP+FP},
\qquad
R=\frac{TP}{TP+FN},
\label{eq:precision_recall}
\end{equation}

where $TP$, $FP$, and $FN$ indicate true positives, false positives, and false negatives, respectively. A detection is treated as correct when its IoU with the assigned ground-truth box is at least 0.5.

\subsubsection{Recognition Metrics}

Character recognition is measured using normalized character accuracy and full-sequence accuracy. Character accuracy is based on the Levenshtein edit distance between the predicted and reference strings:

\begin{equation}
\mathrm{CharAcc}
=
1-
\frac{
\mathrm{Lev}
\left(
T_{\mathrm{pred}},
T_{\mathrm{gt}}
\right)
}{
\max
\left(
|T_{\mathrm{gt}}|,1
\right)
}.
\label{eq:character_accuracy}
\end{equation}

Full-sequence accuracy measures the fraction of plates for which the complete predicted string matches the reference transcription:

\begin{equation}
\mathrm{SeqAcc}
=
\frac{1}{N}
\sum_{i=1}^{N}
\mathbb{I}
\left[
T_{\mathrm{pred}}^{(i)}
=
T_{\mathrm{gt}}^{(i)}
\right],
\label{eq:sequence_accuracy}
\end{equation}

where $N$ is the number of evaluated plate instances and $\mathbb{I}[\cdot]$ is the indicator function.

\subsubsection{Edge Metrics}

Runtime measurements include end-to-end latency, throughput, allocated inference memory, and average device power. End-to-end latency is measured from the start of frame preparation to the completion of bounding-box and text decoding. Throughput is calculated from continuous batch-one execution rather than from isolated engine timing.

\subsection{Baseline Configurations}

The proposed framework is compared with the following ALPR configurations under the same test partition:

\begin{itemize}
    \item \textbf{YOLOv8s + CRNN:}
    A two-stage pipeline in which YOLOv8s localizes the plate and an independent convolutional recurrent neural network processes a resized RGB crop.

    \item \textbf{YOLOv8s + PARSeq:}
    A two-stage architecture combining YOLOv8s localization with a transformer-based sequence-recognition model.

    \item \textbf{YOLO-LPR without RoIAlign:}
    A unified variant that extracts plate features through discrete feature cropping and resizing rather than bilinear RoIAlign.

    \item \textbf{Proposed framework:}
    The shared CSP-Darknet and PANet network with anchor-free localization, feature-space RoIAlign, and CTC sequence recognition.
\end{itemize}

All accuracy measurements were calculated on the held-out test set of 5,550 images.

\begin{table*}[t]
\centering
\caption{Batch-one edge Inference on the Jetson Orin Nano.}
\label{tab:edge_benchmarks}
\begin{tabular}{lccccc}
\hline
\textbf{Configuration}
&
\textbf{Precision}
&
\textbf{E2E Latency (ms)}
&
\textbf{Throughput (FPS)}
&
\textbf{Memory (MB)}
&
\textbf{Power (W)} \\
\hline
YOLOv8s + CRNN
& FP16 & 48.3 & 20.7 & 1,420 & 14.2 \\

YOLOv8s + PARSeq
& FP16 & 112.6 & 8.8 & 2,150 & 14.8 \\
\hline
Proposed framework
& FP32 & 22.4 & 44.6 & 840 & 13.1 \\

Proposed framework
& FP16 & 11.2 & 89.2 & 510 & 11.8 \\

Proposed framework
& INT8 & \textbf{6.8} & \textbf{147.0} & \textbf{315} & \textbf{9.4} \\
\hline
\end{tabular}
\end{table*}

\begin{table*}[t]
\centering
\caption{Effect of plate-feature extraction method.}
\label{tab:ablation_roialign}
\begin{tabular}{lcccc}
\hline
\textbf{Extraction Method}
&
\textbf{Backbone Passes}
&
\textbf{Intermediate Copy (MB/s)}
&
\textbf{Latency (ms)}
&
\textbf{Seq. Acc. (\%)} \\
\hline
RGB image crop
& 2 & 412.5 & 18.4 & 91.2 \\

Discrete RoIPool
& 1 & 0.0 & \textbf{5.1} & 84.6 \\

Bilinear RoIAlign
& 1 & 0.0 & 5.6 & \textbf{91.5} \\
\hline
\end{tabular}
\end{table*}

\subsection{Detection and Recognition Results}

Table~\ref{tab:accuracy_comparison} compares localization and transcription accuracy. The FP32 version of the proposed framework reaches an $\mathrm{mAP}_{50}$ of 98.1\%, a character accuracy of 96.3\%, and a full-sequence accuracy of 91.5\%. INT8 conversion reduces full-sequence accuracy by 0.8 percentage points while retaining an $\mathrm{mAP}_{50}$ of 97.8\%.

The YOLOv8s + PARSeq baseline obtains the highest sequence accuracy at 92.1\%. The proposed FP32 model is 0.6 percentage points lower, while its INT8 version is 1.4 percentage points lower. The unified model avoids the second visual backbone required by both two-stage baselines.

The effect of aligned feature sampling can also be observed by comparing the proposed FP32 model with the YOLO-LPR variant. Adding RoIAlign increases full-sequence accuracy from 83.7\% to 91.5\%, a gain of 7.8 percentage points. This result indicates that bilinear feature sampling is better suited to small plate regions than discrete feature cropping.

\subsection{Edge Deployment Results}

Table~\ref{tab:edge_benchmarks} shows batch-one measurements on the Jetson Orin Nano. The proposed INT8 engine shows the lowest latency and allocated inference memory and the highest throughput among the evaluated configurations.

\begin{figure}[t]
    \centering
    \includegraphics[width=0.95\columnwidth]{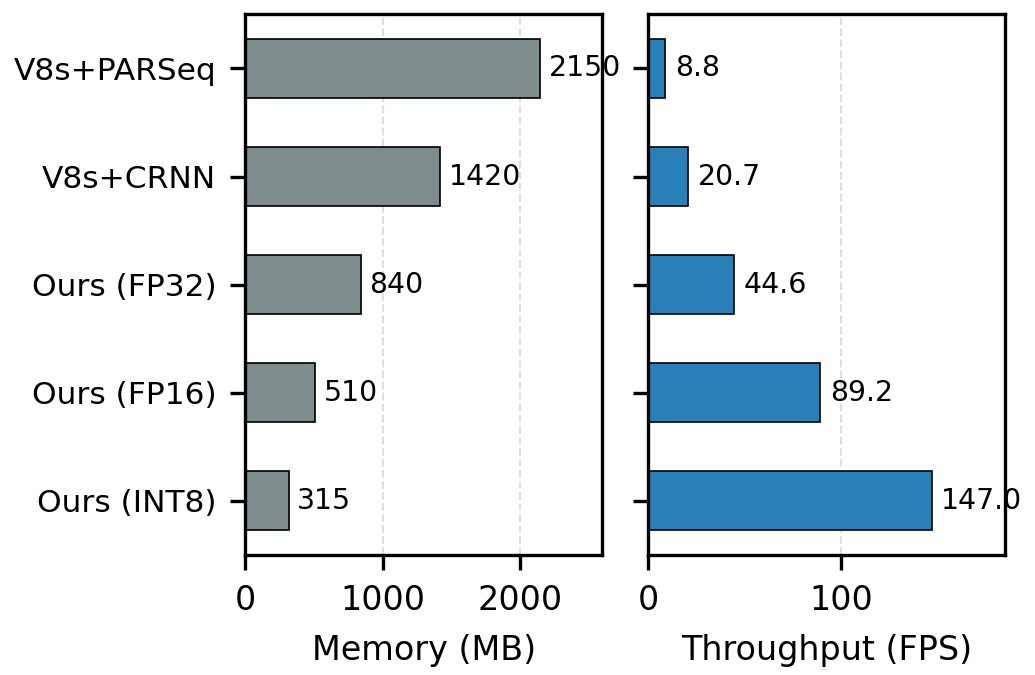}
    \caption{Allocated inference memory and batch-one throughput across the evaluated ALPR configurations.}
    \label{fig:edge_memory_throughput}
\end{figure}

Relative to YOLOv8s+CRNN, the proposed INT8 configuration reduces end-to-end latency from 48.3~ms to 6.8~ms and allocated inference memory from 1,420~MB to 315~MB. Throughput increases from 20.7~FPS to 147.0~FPS, corresponding to a 7.1$\times$ increase and a 77.8\% reduction in allocated inference memory. Figure~\ref{fig:edge_memory_throughput} further shows the progression from FP32 to FP16 and INT8, where throughput increases from 44.6 to 89.2 and 147.0~FPS while memory decreases from 840 to 510 and 315~MB, respectively.

\subsubsection{Latency Breakdown}

\begin{figure}[!ht]
    \centering
    \includegraphics[width=0.88\columnwidth]{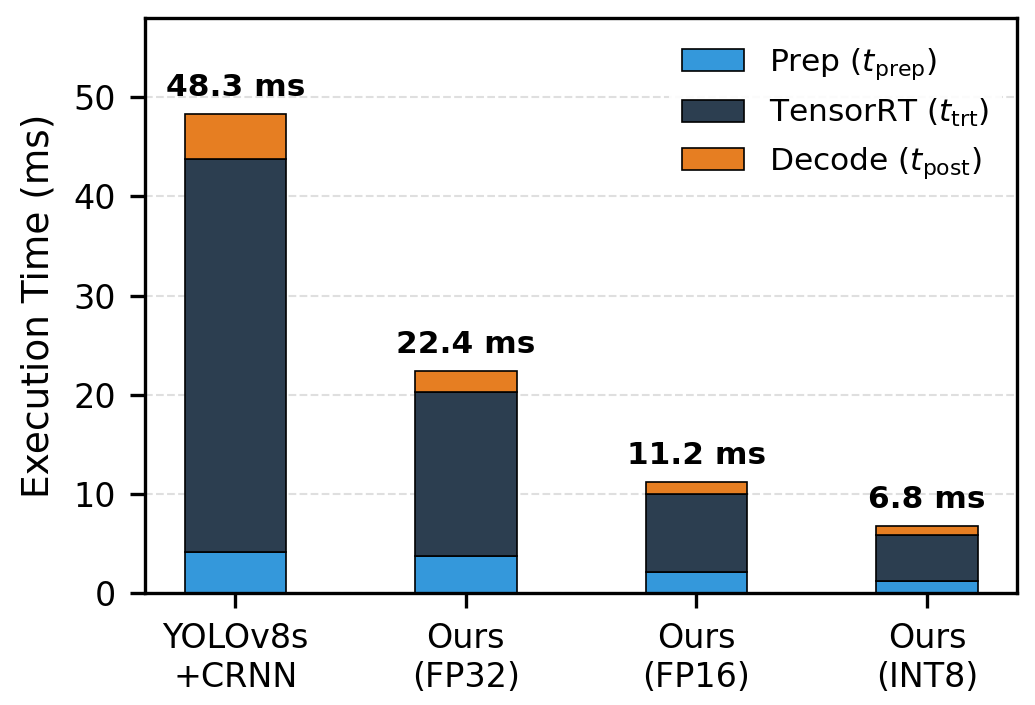}
    \caption{End-to-end latency decomposition of the proposed INT8 deployment into frame preparation, model execution, and output decoding.}
    \label{fig:latency_breakdown}
\end{figure}

As shown in Fig.~\ref{fig:latency_breakdown}, the 6.8~ms end-to-end latency consists of 1.2~ms for frame preparation, 4.7~ms for TensorRT execution, and 0.9~ms for output decoding. TensorRT execution accounts for 69.1\% of the total latency, while preparation and decoding account for 17.6\% and 13.2\%, respectively. The measured runtime gains result from lower numerical precision, shared feature extraction, fixed runtime allocation, and removal of the second recognition backbone. Their individual effects are examined where possible through the following ablation studies.

\subsection{Ablation Studies}

\subsubsection{Feature Extraction Method}

Table~\ref{tab:ablation_roialign} examines the effect of image-space cropping, discrete RoIPool, and bilinear RoIAlign. The image-space configuration evaluates a second recognition backbone, while both feature-space configurations reuse the PANet output.

Discrete RoIPool produces the lowest latency but reduces sequence accuracy by 6.9 percentage points relative to RoIAlign. The loss is associated with coordinate rounding during discrete pooling, which is more harmful when a plate occupies a small region of the feature map. RoIAlign avoids this rounding through bilinear sampling and reaches nearly the same sequence accuracy as image-space recognition while requiring only one backbone pass.

\begin{table}[!ht]
\centering
\caption{Effect of numerical precision and INT8 calibration.}
\label{tab:ablation_quantization}
\begin{tabular}{lcccc}
\hline
\textbf{Precision}
&
\textbf{Calibration}
&
\textbf{Images}
&
\textbf{$\mathrm{mAP}_{50}$ (\%)}
&
\textbf{Seq. Acc. (\%)} \\
\hline
FP32
& Not required & -- & 98.1 & 91.5 \\

FP16
& Not required & -- & 98.1 & 91.4 \\
\hline
INT8
& MinMax & 100 & 94.2 & 78.3 \\

INT8
& Entropy & 100 & 96.5 & 86.1 \\

INT8
& Entropy & 1,000 & \textbf{97.8} & \textbf{90.7} \\
\hline
\end{tabular}
\end{table}

\subsubsection{INT8 Calibration} 
Table~\ref{tab:ablation_quantization} shows calibration method and sample size affect recognition accuracy. MinMax calibration with 100 images reduces sequence accuracy by 13.2 percentage points relative to FP32. Entropy calibration with 1,000 images limits this reduction to 0.8 percentage points, indicating that the calibration subset should include variation in plate layout, scale, lighting, viewpoint, and image quality.

Overall, the results show that the shared network approaches the accuracy of the stronger two-stage baseline while removing its second feature encoder. Bilinear RoIAlign preserves character information better than discrete pooling, and a varied calibration subset is required to retain both localization and sequence-recognition accuracy after INT8 conversion. The resulting INT8 model trades 0.8 percentage points of sequence accuracy for substantially lower latency and inference-memory use on the Jetson Orin Nano.

\section{Conclusion and Future Work}
\label{sec:conclusion}

This paper proposed a unified multi-task framework for license plate detection and character recognition on the Jetson Orin Nano. The network couples a shared CSP-Darknet and PANet feature encoder with an anchor-free localization head and a RoIAlign-based sequence recognition branch. This proposed approach overcomes the requirement of second visual backbone used in conventional two-stage systems and enables recognition directly from aligned feature maps. The FP32 model achieved an $\mathrm{mAP}_{50}$ of 98.1\% and a full-sequence accuracy of 91.5\%. After INT8 model conversion and optimization, the model retained an $\mathrm{mAP}_{50}$ of 97.8\% and a sequence accuracy of 90.7\%. The INT8 engine shows 6.8~ms end-to-end latency, 147~FPS throughput, 315~MB memory allocation, and 9.4~W average power consumption on the target platform. A limitation of this proposed research is that the model was trained and tested primarily on AI-assisted images. Its performance on real roadside images or videos may be affected by sensor noise, uncommon plate formats, severe occlusion, and lighting conditions, which is not covered by the generated dataset. Future work will therefore evaluate the framework on public and real traffic-camera datasets. Further study will also examine mixed synthetic-real training, quantization-aware training, and branch-specific calibration to reduce the remaining INT8 recognition loss.

\bibliographystyle{IEEEtran}
\bibliography{ref.bib}

\vspace{12pt}

\end{document}